\documentclass[conference]{IEEEtran}
\IEEEoverridecommandlockouts
\usepackage{multirow}
\usepackage{cite}
\usepackage{amsmath,amssymb,amsfonts}
\usepackage{algorithmic}
\usepackage{graphicx}
\usepackage{textcomp}
\usepackage{xcolor}
\def\BibTeX{{\rm B\kern-.05em{\sc i\kern-.025em b}\kern-.08em
    T\kern-.1667em\lower.7ex\hbox{E}\kern-.125emX}}
\IEEEpubid{\makebox[\columnwidth]{978-1-7281-6251-5/20/\$31.00~\copyright~2020 IEEE\hfill} \hspace{\columnsep}\makebox[\columnwidth]{ }}

\begin{document}

\title{All-Clear Flare Prediction Using Interval-based Time Series Classifiers\\
}

\author{
    \IEEEauthorblockN{
        Anli Ji\IEEEauthorrefmark{1}\IEEEauthorrefmark{2},
        Berkay Aydin\IEEEauthorrefmark{1}\IEEEauthorrefmark{3}, 
        Manolis K. Georgoulis\IEEEauthorrefmark{4}, 
        Rafal Angryk\IEEEauthorrefmark{5}
        }\\
    \IEEEauthorblockA{
        Department of Computer Science, Georgia State University, Atlanta, GA\IEEEauthorrefmark{1}\IEEEauthorrefmark{2}\IEEEauthorrefmark{3}\IEEEauthorrefmark{4}\IEEEauthorrefmark{5}\\
        Email: aji1@student.gsu.edu\IEEEauthorrefmark{1}\IEEEauthorrefmark{2},
        baydin2@cs.gsu.edu\IEEEauthorrefmark{1}\IEEEauthorrefmark{3},\\
        manolis.georgoulis@phy-astr.gsu.edu\IEEEauthorrefmark{4},
        rangryk@cs.gsu.edu\IEEEauthorrefmark{5}
        }
    }

\maketitle

\begin{abstract}
An all-clear flare prediction is a type of solar flare forecasting that puts more emphasis on predicting non-flaring instances (often relatively small flares and flare quiet regions) with high precision while still maintaining valuable predictive results. While many flare prediction studies do not address this problem directly, all-clear predictions can be useful in operational context. However, in all-clear predictions, finding the right balance between avoiding false negatives (misses) and reducing the false positives (false alarms) is often challenging. Our study focuses on training and testing a set of interval-based time series classifiers named Time Series Forest (TSF). These classifiers will be used towards building an all-clear flare prediction system by utilizing multivariate time series data. Throughout this paper, we demonstrate our data collection, predictive model building and evaluation processes, and compare our time series classification models with baselines using our benchmark datasets. Our results show that time series classifiers provide better forecasting results in terms of skill scores, precision and recall metrics, and they can be further improved for more precise all-clear forecasts by tuning model hyperparameters.

\end{abstract}


\section{Introduction}
Solar flares are one of the manifestations of solar activity that burst out electromagnetic radiation. In the case of large ones, they are often associated with an eruption, in the form of a coronal mass ejection (CME). High-frequency electromagnetic radiation and particles from flares with the associated eruptions can be filtered out by Earth’s atmosphere. However, they still pose a hazard to astronauts and sensitive electronic equipment in space. Additionally, a strong enough CME can induce currents in the Earth’s atmosphere and large networks of power grids. With a flare prediction system, we develop the ability to forecast and send out warning signals prior to the flare event. Over the past three decades, scientists have focused on implementing different model-driven techniques \cite{priest2002magnetic}\cite{shibata2011solar}\cite{Kusano2020} for predicting solar flares based on their intensity. More recently, machine learning-based methods for flare prediction have appeared. A short list of examples including support vector machines (SVMs) in  \cite{Qahwaji2007}, \cite{Bobra2015}, \cite{Boucheron2015}, \cite{Bobra2016}, logistic regression \cite{Song2008}, or decision trees \cite{Yu2009} \cite{Huang2010} among many others.

Solar flare predictions, when modeled as binary classification problem, have flaring and non-flaring class labels. While variations exist, flaring labels are usually associated with occurrence of major flares (flares of class $\ge$M1.0 or $\ge$ X1.0 of the NOAA/GOES flare classification) \cite{Benvenuto2018}. Moreover, besides predicting one or more flare classes, a cumulative flare-index prediction can be attempted \cite{Abramenko2005}. On the other hand, all-clear flare prediction focuses on forecasting the non-flaring class more precisely instead of simply predicting a binary or probabilistic assessment of whether a flare will occur. A predicted all-clear signal indicates that during a prediction time interval (i.e., forecast horizon), no major flares will occur and it is safe (or safer) to operate. Issuing an inaccurate all-clear forecast can be crucial and devastating considering the impact of a missed flare (which corresponds to a false negative) that can cause disruption in the services of many sectors, or even jeopardize an astronaut’s life or health. 
Most scientific studies concentrate on predicting the flare occurrences, which is understandable considering the research aspect of these studies.
Most of the solar flare prediction algorithms implement point-in-time measurements which consist of multiple physical parameters observations with only one individual value for each solar flare event \cite{Boucheron2015}. We envision that the first step towards building operationally-driven, reliable space weather forecasting systems is building low-risk all-clear models. Therefore, our work puts more emphasis on the feasibility of multivariate time series data analysis utilizing an interval-based algorithm named Time Series Forest (TSF) as the base. As a side note, notice the study of \cite{Florios2018} that performs point-in-time forecasting and concludes that random forests are a viable method for flare prediction. Time Series Forest implements a highly specialized random forest and relies on a number of statistical descriptive features such as the mean, standard deviation, and slope of each interval to feed in an ensemble of different decision trees. As for the multivariate time series data, we take advantage of our solar flare benchmark dataset, Space Weather Analytics for Solar Flares (SWAN-SF) \cite{Angryk2020}, which consists of solar photospheric vector magnetograms in Spaceweather HMI Active Region Patch (SHARP) series. 

The rest of the paper is organized as follows. In Section \ref{related_work}, we discuss some related work on well known flare prediction with a focus on all-clear systems. In Section \ref{methodology}, we explain our research methodology on data collection and preparation as well as training and evaluating the interval-based TSF classifiers. In Section \ref{evaluation}, we present our experimental evaluation for our all-clear flare prediction models. In Section \ref{conclusion}, we provide a summary of our findings and discuss our future work avenues.

\section{Related Work}\label{related_work}

Spaceweather HMI Active Region Patches (SHARPs) data product are created using magnetograms generated by the Helioseisemic and Magnetic Imager (HMI) onboard the Solar Dynamics Observatory (SDO)\cite{Bobra2014}, and this product consists of geometric representations of tracked active region magnetogram patches as well as space weather-related parameters from solar photospheric vector magnetic fields. As solar flares are phenomena caused by sudden, abrupt changes in magnetic field in the solar atmosphere, it is well reasoned to build predictive capabilities employing magnetic field parameters \cite{Qahwaji2007}. One of the earlier examples is \cite{Bobra2015}, where Bobra and Couvidat used magnetic field parameters in SHARP data series to forecast M- and X-class flares using SVMs \cite{Bobra2015}. Since then, many researchers started taking advantage of the same idea and implemented a number of different predictive models with various focuses. 

There are also a few studies for all-clear space weather forecasting, particularly for flares and solar energetic particle (SEP) events. Engell et al. \cite{Engell2017} proposed an idea of combining pre-eruptive and post-eruptive forecasts together for an optimization of all-clear forecasts. During the ``All-Clear" forecasts workshop in \cite{Barnes2016}, a systematic comparison between different methods was presented with a recommendation of using temporal features and time series analysis for better forecasting. Other flare prediction algorithms were also demonstrated with the focus on predicting ``all-clear periods'' from the view of solar flares and SEP events \cite{Swalwell2017}. 

In the field of time series analysis, the general category of interval based time series classification algorithms take advantage of descriptive features (often in the form of statistical features) that are derived from fixed or random intervals in each time series. For each time series with length of $n$, there exists $n(n-1)/2$ possible contiguous intervals. Each of the temporal features that are calculated over a certain time series interval have the ability to capture some important characteristics of the series. However, generating features from all possible intervals is not feasible as a dataset with a large number of parameters will lead to the interval feature space to grow exponentially. Deng et al. \cite{Deng2013} proposed an implementation of a random forest approach using only calculated statistical values of each interval as features to overcome this exponential feature space problem. This interval-based random forest (ensemble) approach is the TSF where multiple decision trees are grouped together. Each tree in this ensemble is trained using a subset of statistical features derived from randomly selected intervals, which plays an essential role in reducing high feature spaces. These features include simple and effective mean, standard deviation and slope extracted from those intervals. In Section \ref{methodology}, we provide more details about these statistical values and the classifier itself.

Our work implements an interval-based classifier for solar flare prediction via a multivariate time series forest. This will not only provide a multivariate schema to extend the univariate time series classifier but also build a prototype all-clear flare prediction system which optimizes the model based on forecast skill scores. We hope this work affords researchers and practitioners in related fields more exposure to multivariate time series analysis for space weather forecasting.

\section{Methodology}\label{methodology}
\subsection{Data Collection}

\textbf{S}pace \textbf{W}eather \textbf{AN}alytics dataset for \textbf{S}olar \textbf{F}lare prediction (SWAN-SF) is a recently introduced dataset by Angryk et al. \cite{Angryk2020}. It is an open source multivariate time series (MVTS) dataset (can be accessed from Harvard Dataverse \cite{swansf}) that provides time series data for a collection of 24 space weather-related physical parameters primarily calculated from magnetograms. The time series parameters examined in this research are Schrijver's R-value \cite{Schrijver2007} and total unsigned flux \cite{Leka2003} (R\_VALUE and USFLUX in HARP keywords). For evaluating the performance of our proof-of-concept multivariate time series classifiers, we will use these two parameters, although the analysis can be extended to include all 24 parameters. 

The SWAN-SF dataset consists of five partitions that cover the period of May 2010 to August 2018. Each of these partitions contains approximately an equal sum of large flares (i.e., X- and M-class). These partitions are time-segmented which means the data instances across partitions do not have a time overlap. The active regions are sliced with a sliding observation window for both flare-quiet regions as well as each verified flare with a corresponding active region number (in this case NOAA Active Region number matched to HARPNUM, unique HARP identifier for an active region patch series, keyword based on \cite{cai2019application}),

A sliding observation window (a 12 hour interval) iterates over the multivariate time series (with 1 hour step size) and checks if there exists a set of flares associated to the active region in the next 24 hours. Each slice is then labeled as the maximum intensity flare originating from that active region. In the case when no flares are present, a flare-quiet label is used. Note here that major flaring categories (X, M, C, B, or A) are based on logarithmic classification of peak X-ray flux. We use instances labeled with M- and X-class flares as our flaring class in this work. The remaining instances with relatively weak C- and B- class flare labels and flare-quiet ones are considered as non-flaring class. We will refer to flaring class as `positive' class and non-flaring as `negative'.  Doing that, we model the flare forecasting problem as a binary multivariate time series classification task. 

\begin{figure}[t]
    \centering\includegraphics[width=\linewidth]{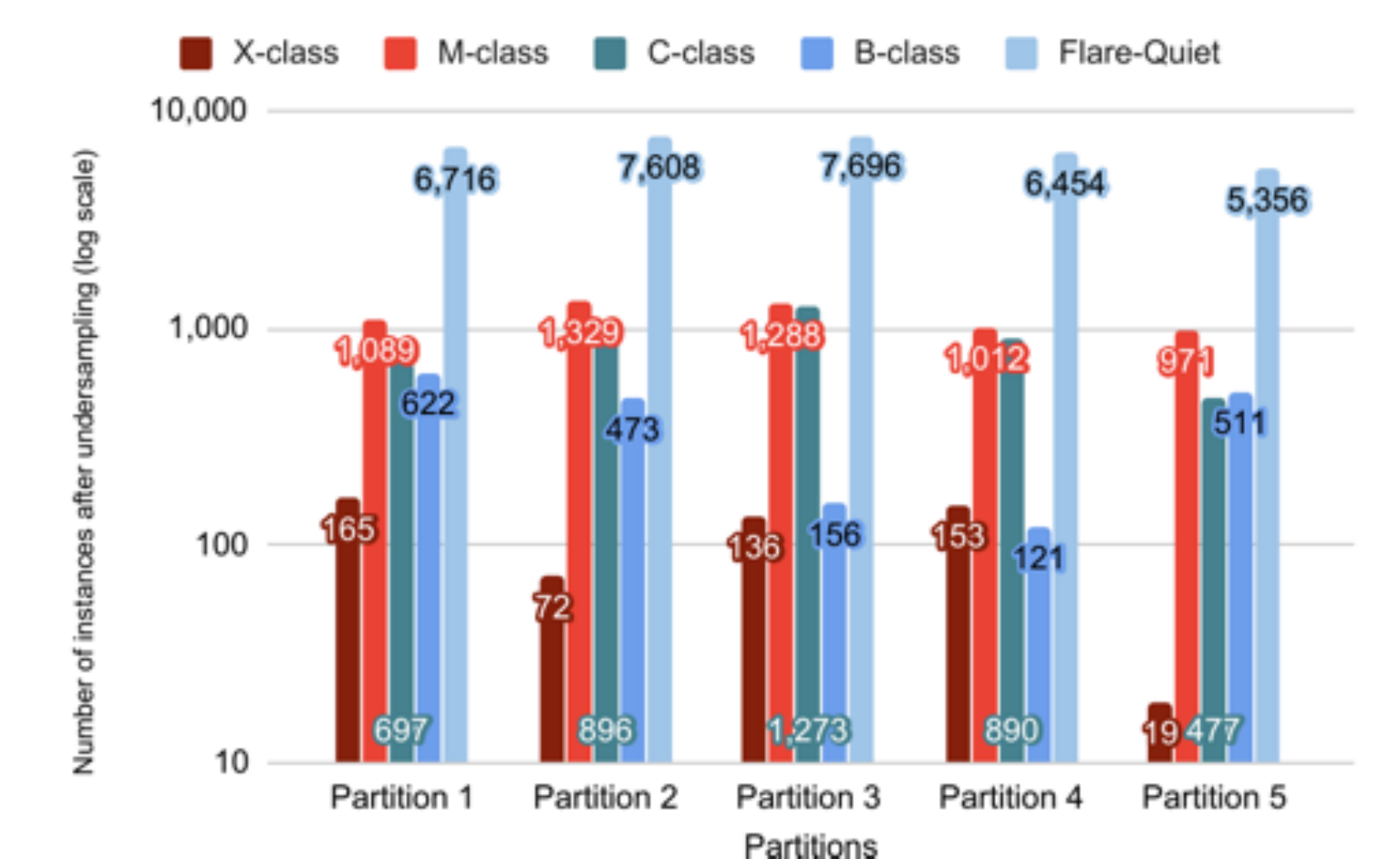}
    \caption{The number of instances for X-, M-, C-, and B-class flare or flare-quiet labels in each partition for the climatology-based undersampling scenario. The partitions are time segmented. Each partition has similar total numbers for M- and X-class flares. 
    }
    \label{fig:clus}
\end{figure}

In our early stages of study, we have also used an undersampled dataset derived from SWAN-SF. This undersampled version of the dataset reduces the number of C-, B- class and flare-quiet instances in the non-flaring (which is the majority class) based on the daily climatology estimates to individual active region, eventually reducing the overall class imbalance ratio from $\sim$1:50 to $\sim$1:6.4. We provide the number of multivariate time series data instances per different flare-classes for five time-segmented partitions in Figure~\ref{fig:clus}. Details of climatology-undersampled dataset (which we will refer to as CLUS, hereafter) generation can be found in \cite{Ahmadzadeh2019}. The CLUS dataset, also partitioned using time-segmented partitioning, covers the same active region multivariate time series as the original SWAN-SF, and is created with 12-hour observation, 24-hour prediction windows. CLUS dataset has an advantage over the original SWAN-SF dataset in that it preserves less non-flaring (i.e., C-, B- class, and flare-quiet) instances. As mentioned before, each partition stores an identical or similar number of large flare ($\geq$M1.0) instances. Even though the CLUS dataset is still not balanced
its imbalance is much less extreme than the near-operational case in which all instances from original SWAN-SF dataset is used.
Meanwhile, we believe it is worth mentioning that although the provisional usage of the CLUS dataset in testing allows us to perform time-efficient comparisons in the large set of experiments we performed, it does not represent the actual or expected performance of the models in an operational environment.

\subsection{Methods}
\begin{figure}[b]
    \centering\includegraphics[width=\linewidth]{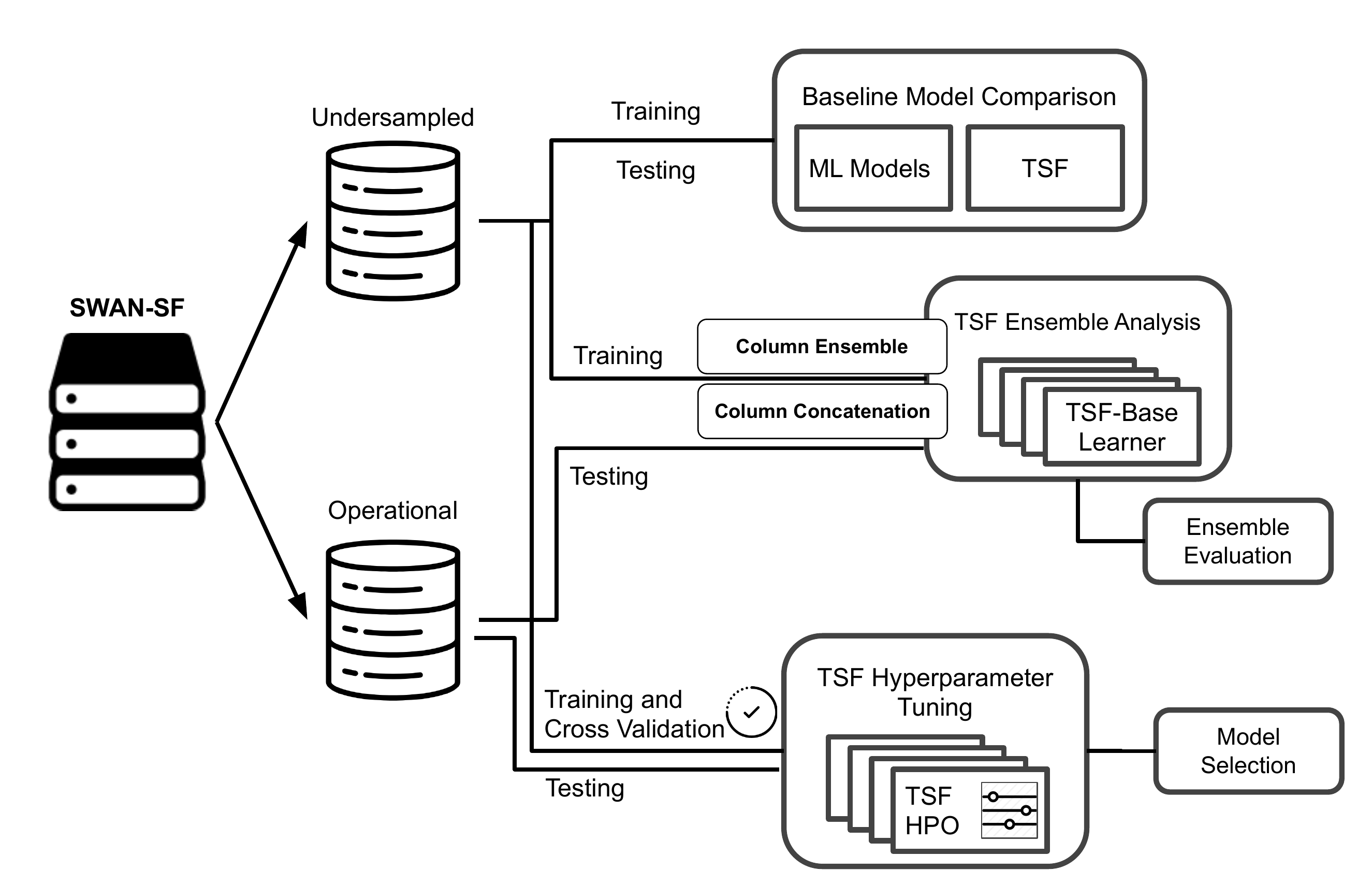}
    \caption{Schematic overview of the homogeneous ensemble pipeline
    }
    \label{fig:TSFschema}
\end{figure}

TSF is originally designated as a univariate time series classifier where it makes use of a single parameter and builds a random forest from statistical features derived from random intervals. In our case of multivariate time series from active region patches, we create a multivariate variant of this algorithm. Specifically, all the multivariate parameters are being handled with different techniques. One way of handling multivariate data is by \emph{column concatenation}, where all the time series columns appended together into one single long time series column. This long column will then be fed into a single univariate TSF classifier to train it. Another method for building a multivariate time series classifier is by the \emph{column ensemble} method. This is a parameter-wise ensemble method of columns in which every parameter (column) will have one classifier fitted. The prediction results that come out from these individual column classifiers will then be aggregated as a whole (with equal vote using prediction probabilities). This is a homogeneous ensemble schema and overview of it is shown in Figure~\ref{fig:TSFschema}.

Another important process in applying machine learning algorithm is hyperparameter tuning. Hyperparameters are values defined before the training process. Hyperparameter optimization is where the optimal hyperparameter sets are selected for the best of performance of a classifier. Grid search is an exhaustive search designed for optimizing hyperparameter settings over all the parameter combinations of an estimator. Every combination of a predefined hyperparameter setting will be placed into the searching process. This process is applied to determine the optimal hyperparameters of our TSF classifier. Alternatively, grid search with cross validation (GridSearch CV) \cite{sklearn_api} is analogous to grid search where the searching classifier in this case performs an out-of-sample validation for further testing the model’s predicting ability. Indeed, two subdivided training and testing sets are extracted from the input training data to serve as cross validation subsets to derive a more accurate estimation of the model predicting performance. 

A traditional grid search cross-validation (CV) schema is targeted for tabulated data and assumes that the instances are independent, meaning the random assignments of instances to different training and testing folds (or partitions) do not potentially create an overfitting or a memorization issue for trained models. For time series data, sampled with a sliding window, portions of the time series will be repeated. This will lead to fairly similar instances put into both training and testing sets, and the creation of models with a tendency to memorize, rather than learn. While these models may potentially provide better initial results, it would not be due to their stronger generalization capabilities but a sub-optimal sampling choice \cite{Ahmadzadeh2019}. In time series analysis, where instances are obtained with a sliding window---creating overlapping time series instances, the data partitions for training, testing and validation are required to be time segmented. With the use of a traditional grid search CV schema, we cannot guarantee that the data instances coming from consecutive overlapping segments are not placed both in training and testing sets, which will be detrimental to the reliability of time series classification performance evaluations. To alleviate this issue, we implemented a customized CV schema which modified the original grid search to split by the SWAN-SF (or CLUS) partitions for the purpose of maintaining continuous time segmentation instead of randomized sampling. Each time segmented training partition dataset is assigned with a partition label. For data instances with the identical label, they will not be placed into both testing and training sets. In addition to the customizing the random grid search, we also modified the scoring function for our CV and replaced the classification accuracy with forecast skill scores, mainly Heidke Skill Score, which we will discuss more in the next section. 

\section{Experimental Evaluation}\label{evaluation}
The experiments are designed for the purpose of fitting a univariate model to a multivariate time series architecture for an all-clear flare prediction. Our aim is to demonstrate the robustness and compare the efficiency of time series classifiers and ensembles for predicting an all-clear signal. All the traditional (i.e., point-in-time, tabulated) classifiers used in this study have been built using Python with scikit-learn library \cite{scikit-learn}. The Time Series Forest classifier that we use for our comparison is from sktime toolkit \cite{https://doi.org/10.5281/zenodo.3749000}. All our source code is open sourced for improving the reproducibility and experiments can be see in our Project Repository \cite{dmlabacfp}.

\subsection{Experimental Settings}
We used the undersampled version of the SWAN-SF dataset (namely the CLUS dataset) as our main sampling techniques. All the missing values in the datasets have been interpolated by using linear interpolation between adjacent data points. For the model evaluation, we implemented a $2\times2$ contingency matrix as for forecasting results in binary (i.e., flare/non-flaring). Based on this matrix, we implemented multiple evaluation metrics as well as essential forecast skill scores. In these metrics measurements, TP (true positives), TN (true negatives), FP (false positives), and FN (false negatives) are used in standard settings, where positive stands for occurrence of a large flare (i.e., X- and M-class [denoted as XM]) while negative stands for comparatively smaller flare and flare-quite region (i.e., C-, B-class and non-flaring/quiet regions [denoted as CBN]). Note here TP is where the model correctly predicts the flare instances (positive class) while TN is where the model correctly predicts the the non-flaring instances (negative class). Both FP and FN represent incorrect results, where FP corresponds to false alarms (non-flaring instances predicted as flare) while FN corresponds to misses (flare instances predicted as non-flaring). Then, probability of false detection (POFD, see Eq.~\ref{eq:POFD}) is the false alarm rate which is calculated as the ratio between FP to all the actual negative class instances.

\begin{equation}
    \label{eq:POFD}
        POFD = \frac{FP}{FP + TN}
\end{equation}

In addition, we used the True Skill Statistic score which compares the difference between the probability of detection (Recall for positive class) and the probability of false detection (POFD). The score measurement is shown in Eq.~\ref{eq:tss}.

\begin{equation}
    \label{eq:tss}
        TSS = \frac{TP}{TP + FN} - \frac{FP}{FP + TN}
\end{equation}

The Heidke Skill Score measures the improvement of the forecast over a random forecast. HSS ranges between -$\infty$ and 1, with 1 indicating perfect performance and 0 indicating no skill. A no skill means that the forecast is not better than a random binary forecast based on class distributions. HSS is given by Eq.~\ref{eq:hss2} where $P\!=\!TP\!+\!FN$ and $N\!=\!FP\!+\!TN$ corresponding to the observed positives and negatives, respectively. 

\begin{equation}
    \label{eq:hss2}
        HSS = \frac{2 \cdot ((TP \cdot TN) - (FN \cdot FP))}{P \cdot (FN + TN) + N \cdot (TP + FP)}
\end{equation}

The Gilbert Skill Score considers the number of hits due to chance, which is given as frequency of an event multiplied by the total number of forecast events. This score formula is given by Eq.~\ref{eq:gss}

\begin{equation}
    \label{eq:gss}
    \begin{split}
        GSS = \frac{TP-CH}{TP+FP+FN-CH} \text{,~~~~~~~~~~~~~~~} \\ 
        \text{~~~~~where } CH = \frac{(TP+FP) \times (TP+FN)}{TP+FP+FN+TN}
    \end{split}
\end{equation}

In addition, we also used the precision and recall scores for both positive and negative classes, encoded as XM and CBN respectively. Their formulas are provided in Eq.s~\ref{eq:precision_xm},\ref{eq:recall_xm},\ref{eq:pre_cbn} and \ref{eq:recall_cbn}. False alarm ratio (FAR) shown in Eq.~\ref{eq:FAR} is the ratio between FP (where predict XM class wrong as CBN) and predicted positive (a total number of XM predictions). 

\begin{equation}
    \label{eq:precision_xm}
        Precision(XM) = \frac{TP}{TP+FP}
\end{equation}

\begin{equation}
    \label{eq:recall_xm}
        Recall(XM) = \frac{TP}{TP+FN}
\end{equation}

\begin{equation}
    \label{eq:pre_cbn}
        Precision(CBN) = \frac{TN}{TN+FN}
\end{equation}
 
\begin{equation}
    \label{eq:recall_cbn}
        Recall(CBN) = \frac{TN}{TN+FP}
\end{equation}

\begin{equation}
    \label{eq:FAR}
        FAR = \frac{FP}{FP+TP}
\end{equation}

We conducted multiple experiments on the TSF classifier using the evaluation metrics mentioned above. In the first experiment, we compared the TSF classifier with simple machine learning classifiers running on tabulated data (see Table 1). For the integrity of our evaluations, all the classifiers included in this experiment are trained with Partitions 1 and 2 from climatology-undersampled (CLUS) dataset and tested with Partition 4 from the original SWAN-SF dataset. As for the second experiment, we implemented a homogeneous ensemble schema where the predictive results of four TSF classifiers trained under different hyperparameter settings are fed into a decision tree meta-learner \cite{cart}. Each of the base TSF learner is trained with Partition 1 from the CLUS dataset and tested with Partition 2 from the same CLUS dataset. The final meta-learner (ensemble model), which is a decision tree classifier, is trained with Partition 3 from the CLUS dataset and tested with Partition 4 from the same dataset. In the third experiment, we generated a hyperparameter tuning procedure by applying our customized grid search with cross validation. For tuning the classifier, the data instances from Partition 1 to 3 in the CLUS dataset is utilized for the training process as well as data from Partition 4 in the original SWAN-SF dataset for the testing process.

\subsection{Comparison with Traditional Machine Learning Models}
In the first experiment, we train multiple models based on well-known machine learning algorithms and compare each of them with results from our TSF classifier. Three simple models are selected: decision tree \cite{cart}, logistic regression \cite{Song2008}, and support vector machine (SVM) \cite{cortes1995support}. Each of the models are trained with five statistical features (i.e., mean, standard deviation, minimum, maximum, and median) from R-value and USFLUX parameters. On the other hand, the TSF classifier itself samples out random intervals from the series and internally evaluates and uses mean, standard deviation and slope of each random interval. These extracted values are concatenated by the classifier to form a new dataset on the fly and builds a random forest model on top. 

\begin{table}[htbp]
\caption{The evaluation metrics for simple classifiers and Time Series Forest trained and tested with undersampled datasets.}  
\renewcommand{\arraystretch}{2.25}
    \setlength{\tabcolsep}{4pt}
    \centering
    \begin{tabular}{r c c c c c c}
    \hline
    Model &
    TSS &
    HSS &
    \begin{minipage}{0.05\textwidth} \centering Precision XM \end{minipage} &
    \begin{minipage}{0.05\textwidth} \centering Precision CBN \end{minipage}  &
    FAR &
    \begin{minipage}{0.05\textwidth} \centering Recall (TPR)  \end{minipage}
     \\ \hline
    \begin{minipage}{0.08\textwidth} \raggedleft Decision Tree \end{minipage} &
      0.69 &
      0.15 &
      0.10 &
      0.996 &
      0.109 &
      0.80 \\ 
    \begin{minipage}{0.07\textwidth} \raggedleft 
    Logistic Regression \end{minipage} &
      0.65 &
      0.17 &
      0.11 &
      0.995 &
      0.089 &
      0.736 \\ 
    \begin{minipage}{0.12\textwidth} \raggedleft 
    Support Vector Machine \end{minipage} &
      0.51 &
      0.17 &
      0.11 &
      0.99 &
      0.067 &
      0.581 \\ 
    \begin{minipage}{0.08\textwidth} \raggedleft 
    Time Series Forest \end{minipage} &
      0.75 &
      0.57 &
      0.50 &
      0.98 &
      0.135 &
      0.885 \\ \hline
    \end{tabular}
\end{table}

The detailed evaluation metrics for the simple classifiers and TSF are shown in Table I. Our results demonstrate that the use of this time series classifier has improved the performance in virtually every aspect other than the precision of non-flaring (CBN) class. Although this is accepted as many of the flaring instances are misclassified by ordinary machine learning techniques. We can suggest that high precision on flaring class (XM) and high recall (TPR) lead to significantly higher results for TSF models. However, it can be noted that the false alarm rate for TSF is actually higher than the others.

\subsection{Impact of Multivariate Ensemble on Model Performance}
As discussed earlier, we build a homogeneous ensemble schema where our base models will employ identical learning algorithms and identical data under different hyperparameter settings. The meta model, which is chosen as the decision tree classifier, will assemble all the results returned from the base models. In Table II, we provide detailed settings for each individual base model. Each pair of the base learners will be using the Column Concatenation and Column Ensemble techniques for handling multivariate data.

\begin{table}[htbp]
\caption{Tabulate classifier vs Time series Forest classifier}  
\resizebox{\columnwidth}{!}{%
    \begin{tabular}{|c|c|c|c|c|c|c|}
    \hline
    \begin{tabular}[c]{@{}c@{}}Base\\ Model\end{tabular} & Model & \begin{tabular}[c]{@{}c@{}}Multivariate\\ Concatenation\end{tabular} & \multicolumn{2}{c|}{Class weight} & 
    \begin{tabular}[c]{@{}c@{}}Number of\\ Estimators \end{tabular}
    & 
    \begin{tabular}[c]{@{}c@{}}Max\\ Depth \end{tabular}
    \\ \cline{4-5} & & & XM & CBN & & \\ \hline
    BaseModel1 & TSF & \begin{tabular}[c]{@{}c@{}}Column\\ Concatenation$^{\mathrm{1}}$\end{tabular} & 0.33 & 0.67 & 50 & 3 \\ \hline
    BaseModel2 & TSF & \begin{tabular}[c]{@{}c@{}}Column\\ Concatenation$^{\mathrm{1}}$\end{tabular} & 0.33 & 0.67 & 250 & 6 \\ \hline
    BaseModel3 & TSF & \begin{tabular}[c]{@{}c@{}}Column\\ Ensemble$^{\mathrm{2}}$\end{tabular} & 0.33 & 0.67 & 50 & 3 \\  \hline
    BaseModel4 & TSF & \begin{tabular}[c]{@{}c@{}}Column\\ Ensemble$^{\mathrm{2}}$\end{tabular} & 0.33 & 0.67 & 250 & 6 \\ \hline
    \end{tabular} %
    
}
\vspace{0.1em}
\newline 
{\scriptsize $^{\mathrm{1}}$ Column Concatenation appends multiple the time series columns into a single long time series column}

{\scriptsize $^{\mathrm{2}}$ Column Ensemble is parameter-wise ensembling of columns in which one classifier is fitted for each time series column and their predictions aggregated. 
}
\end{table}

All-clear forecasts require a distinct attention to precise and sensitive prediction of non-flaring (CBN) class. If we simply issue all predictions as the majority class (CBN), the baseline precision score will be randomized by about 86.5\% respectively. As shown in Figure 4, all of the models we trained have performed significantly better than the baseline statistical random models (that simply predict all non-flaring or randomly assign classes based on their distributions). Moreover, it is desirable to have a high true negative rate (TNR) without sacrificing much from precision and recall for XM classes. As we perform the homogeneous ensemble schema, our meta learner model has slightly increased in the key accuracy performance measures such as precision (for detecting CBN class instances) and recall (for XM class instances). In predictive models with class imbalance, precision and recall are often inversely related, which means models with high precision often have lower recall values and vice versa. Hence, the TNR score and the precision for XM classes are slightly decreased as compared to the base models.

\begin{figure}[t]
    \centering\includegraphics[width=\linewidth]{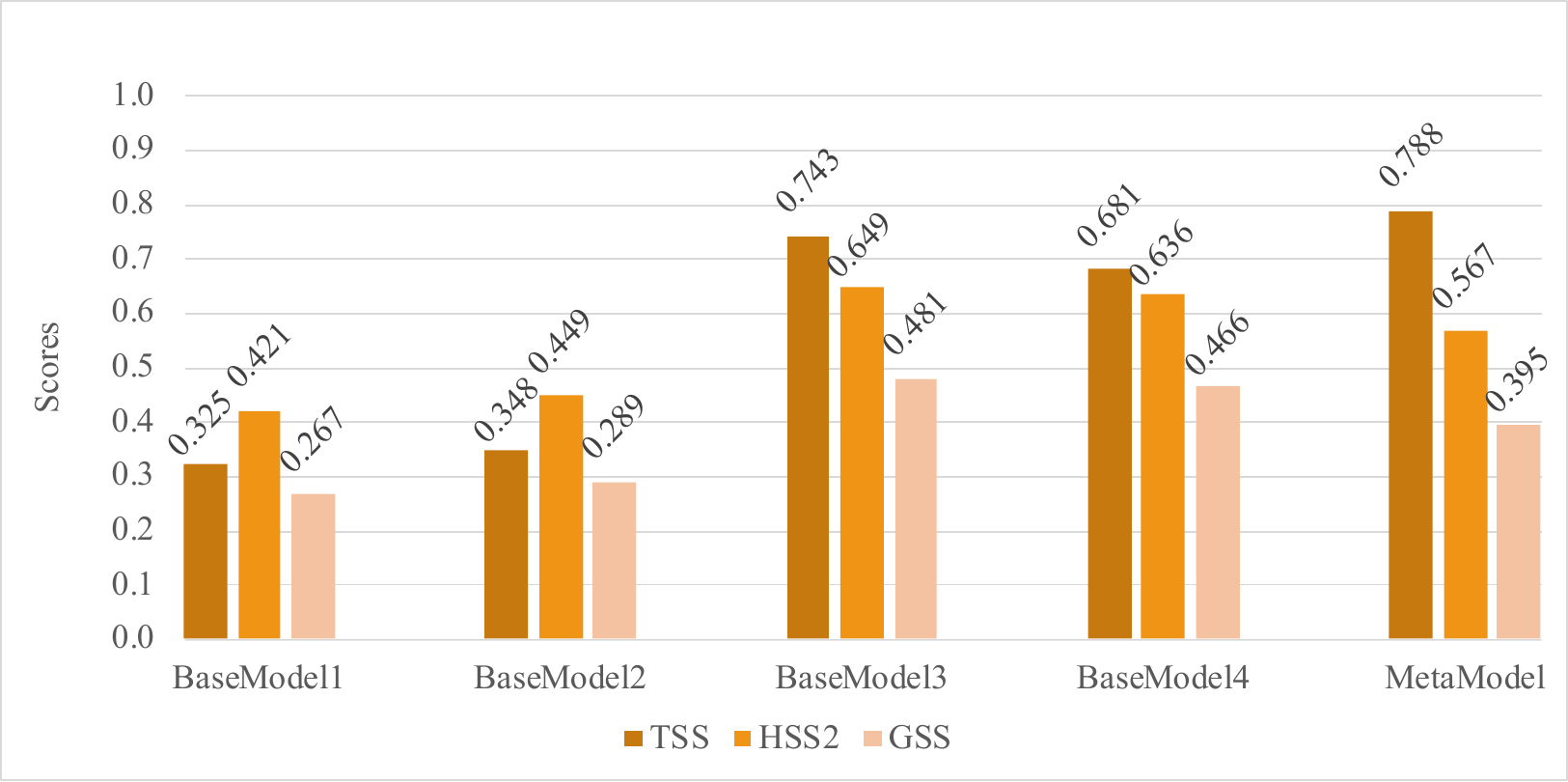}
    \caption{The True Skill Statistic (TSS), Heidke Skill Score (HSS2) and Gilbert Skill Score (GSS) performance evaluation results for Base Learners and Meta-learner models. Results are taken from undersampled datasets (Partition 1 and 2 for Base Learners and Partition 4 for Meta-learner) 
    }
    \label{fig:skillscores}
\end{figure}

\begin{figure}[t]
    \centering\includegraphics[width=\linewidth]{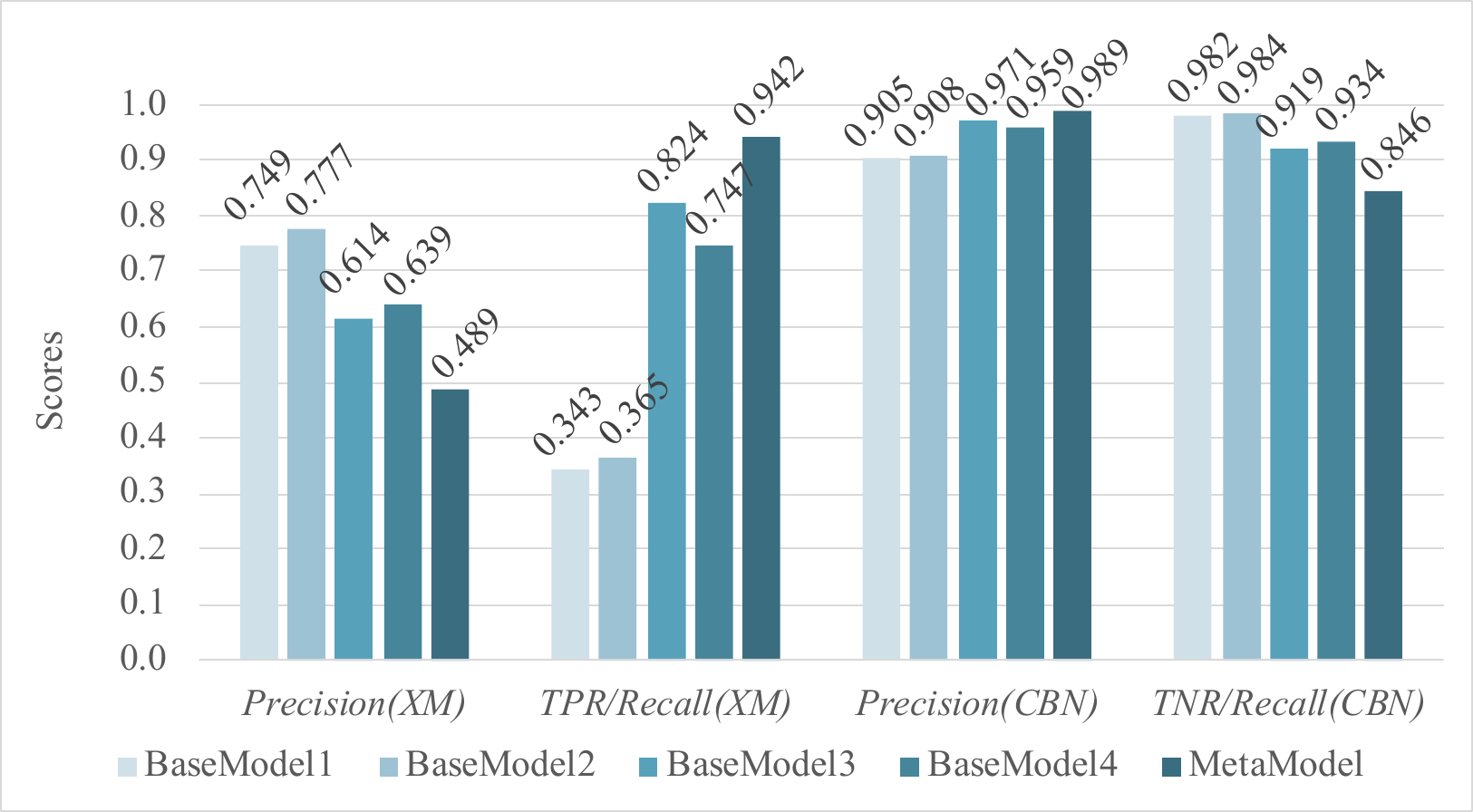}
    \caption{The precision and recall values for both XM and CBN classes for Base Learners and Meta-learner models. Results are taken from undersampled datasets (Partition 1 and 2 for Base Learners and Partition 4 for Meta-learner)
    }
    \label{fig:pre_recall}
\end{figure}

\begin{figure}[t]
    \centering\includegraphics[width=\linewidth]{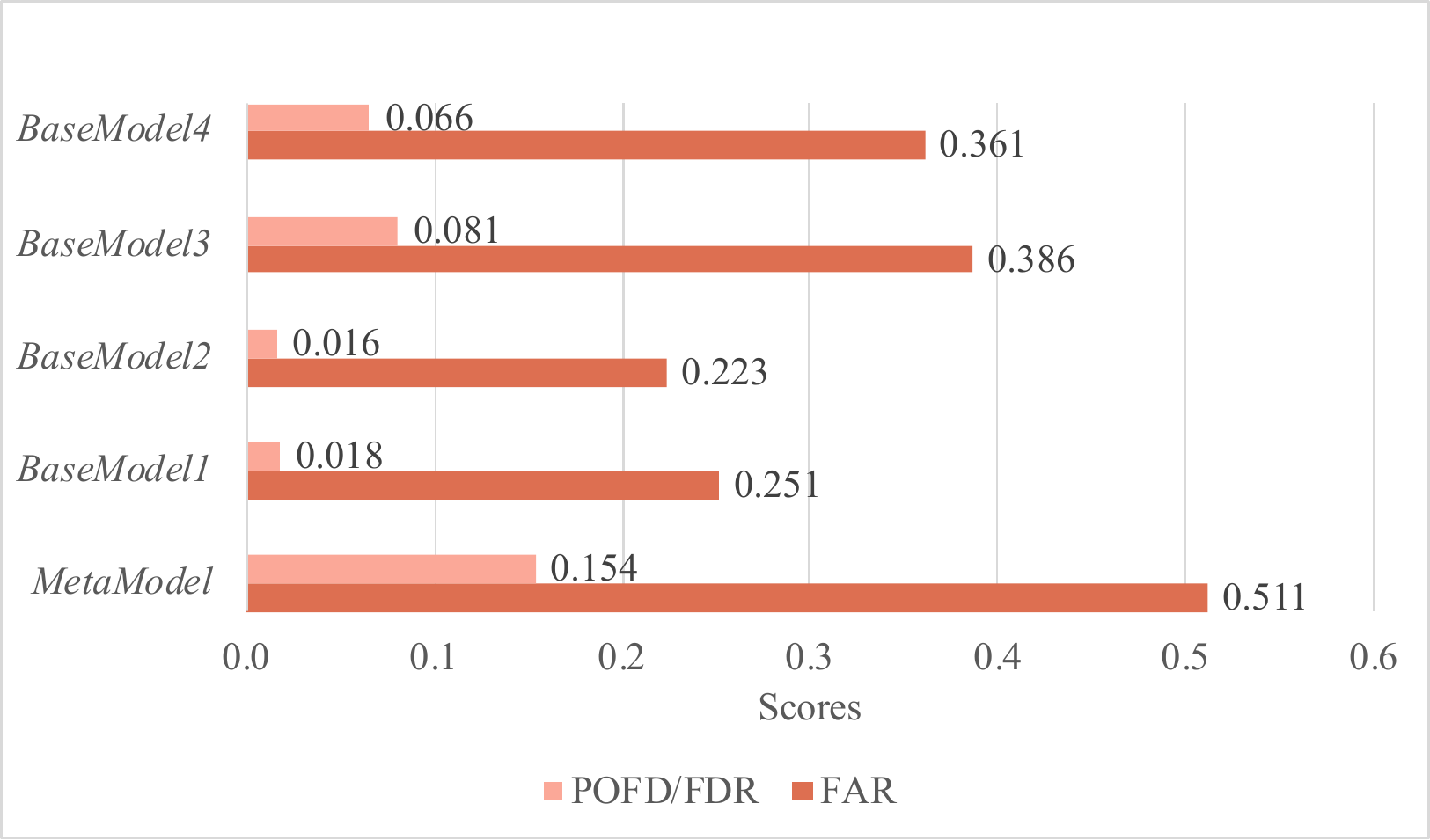}
    \caption{The false alarm ratio (FAR) and probability of false detection (POFD) values (specifically for XM class) for Base Learners and Meta-learner models. Results are taken from undersampled datasets (Partition 2 for Base Learners and Partition 4 for Meta-learner)
    }
    \label{fig:pofd_fpr}
\end{figure}

\begin{figure*}[tbh!]
    \centering
    \includegraphics[width=0.99\textwidth]{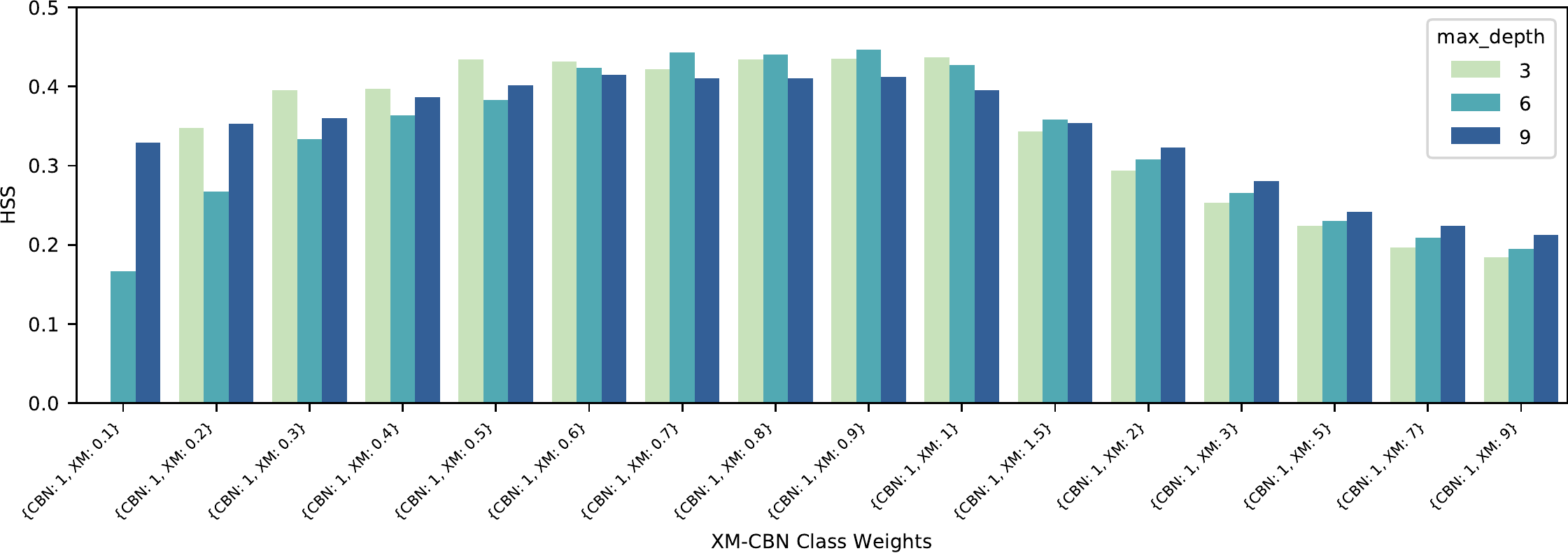}
    \caption{Grid search results based on HSS score with parameters of 100 estimators. Results are driven from training with undersampled datasets (Partitions 1-3 of CLUS) and testing with SWAN-SF datasets (Partitions 4 and 5). Different column colors indicate different maximum depth values.}
\vspace{2em}
    \centering
    \includegraphics[width=0.99\textwidth]{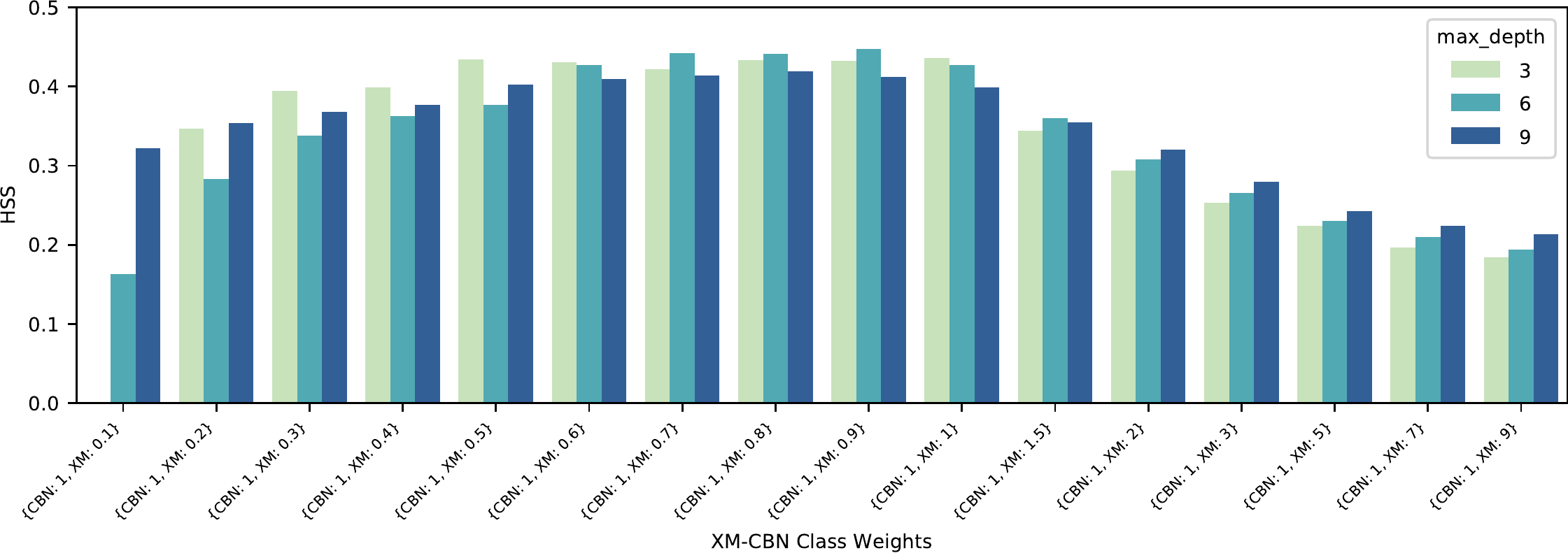}
     \caption{Grid search results based on HSS score with parameters of 500 estimators. Results are driven from training with undersampled datasets (Partition 1-3 of CLUS) and testing with SWAN-SF datasets (Partition 4 and 5). Different column colors indicate different maximum depth values.
     }
     \label{fig:est500}
\vspace{2em}
     \centering
     \includegraphics[width=0.99\textwidth]{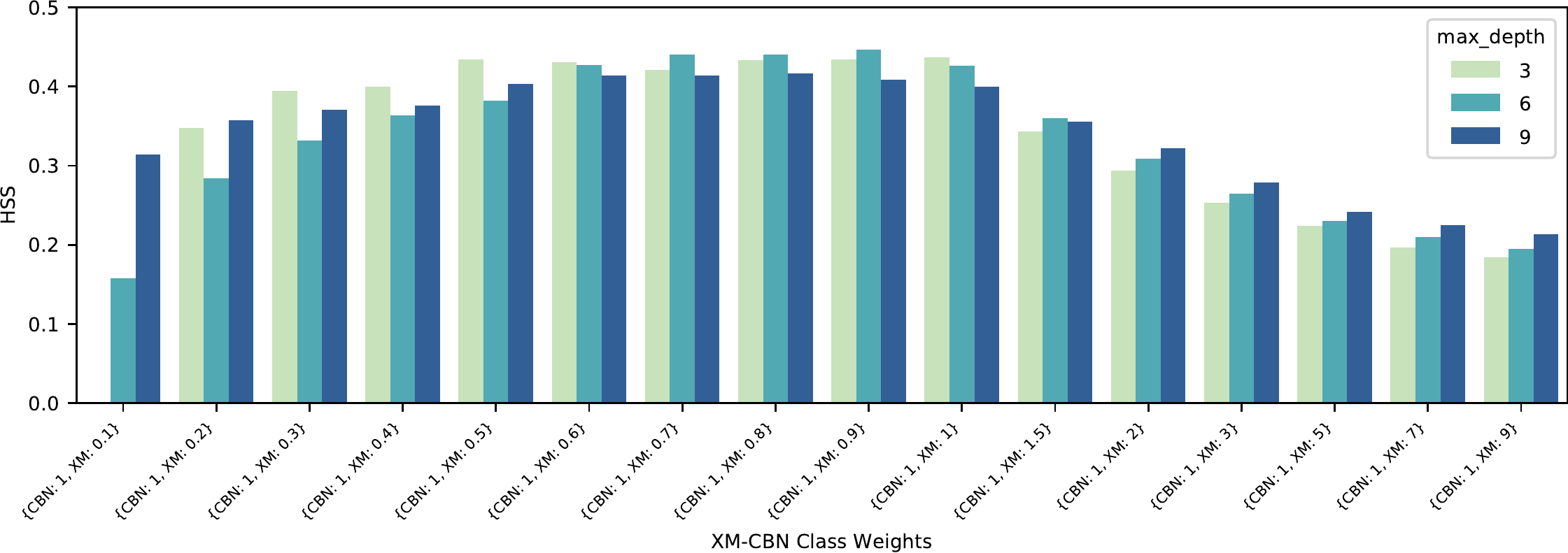}
     \caption{Grid search results based on HSS score with parameters of 1000 estimators. Results are driven from training with undersampled datasets (Partition 1-3 of CLUS) and testing with SWAN-SF datasets (Partition 4 and 5). Different column colors indicate different maximum depth values.
    } \label{fig:est1000}
\end{figure*}

The skill scores shown in Figure 3 compare the model performance that randomly classify all the samples to the majority class. Note here the overall skill of a forecasting system can be minimized if we assign all the predictions to CBN classes, in which it could produce the exact baseline precision scores mentioned earlier. As a result, in Figure 3, even though the TSS of the meta model is slightly increased, the overall skill score of the meta model is relatively lower than both base models 3 and 4. 

The measurement scores of both the false alarm ratio (FAR) and probability of false detection (POFD) show the quality of predicting XM classes. The lower these scores are, the better performance a model has. The approach that aims for maximizing the precision and recall of CBN classes comes with the cost of increasing the false alarm ratio and false detection scores. As shown in Figure 5, both FAR and POFD scores are higher for the meta model despite lower scores being desired.

\subsection{Hyperparameter tuning and class weight adjustments}

As we discussed earlier, there is a high imbalance ratio between our positive and negative classes which will likely benefit accuracy. As the model can simply classify all instances as the majority class, a large accuracy score can be returned as a result. For our all-clear flare forecasting, we would like to see the overall performance of the model in terms of skill and being able to generate high precision and recall forecasts instead of just `naive' accuracy. In this case, we used HSS as our primary scoring function for our hyperparameter optimization although the use of any other scoring function could still work, in principle. In our hyperparameter tuning, we searched up in three dimensions for the number of estimators (number of trees in TSFs), the maximum depth of each estimator, and the class weights between XM and CBN classes.

For the hyperparameter optimization in this experiment, our results provide valuable information (shown in Figure 6-8). In our best results, TSF models with 100 estimators have almost identical HSS value distributions as those models with 500 and 1000 estimators. This is an important observation for decreasing running time requirements for our project. We can also see that the best case for HSS is obtained when the class weights between XM and CBN classes are trending towards 1:1 to 1:0.5 among all settings. Even though the maximum depth of the trees are less important as compared to the number of estimators needed to be trained, this can still allow for slight changes. Larger trees do not always guarantee the best HSS performance. Simpler trees can provide better generalizations and robustness. Meanwhile, these results can still change as optimizing for other metrics. 

\section{Conclusion and Future Work}\label{conclusion}
In this work, we have trained interval-based TSF models for the task of All-Clear flare forecasting. Our models focused on a binary classification schema, from the perspective of all-clear predictions, which concentrates on high precision of non-flaring class predictions and aims to lower the number of misses. Our trained models show that they have better performance than the other machine learning counterparts and they can be further optimized with both hyperparameter tuning of class weights as well as their robustness can be increased with ensemble learners. In our experiments, we have used two time series parameters and even for those two our results are in acceptable levels. This shows that time series classifiers have great potential to improve the all-clear forecasts and is a suitable predictive model for use in operational context. There are more avenues that can be explored for future work which includes but not limited to extending the analysis to other well-known magnetic field parameters, building different ensemble strategies, or optimizing with other evaluation metrics.

\section*{Acknowledgment}

This project has been supported in part by funding from the funding from Division of Advanced Cyberinfrastructure within the Directorate for Computer and Information Science and Engineering, the Division of Astronomical Sciences within the Directorate for Mathematical and Physical Sciences, and the Division of Atmospheric and Geospace Sciences within the Directorate for Geosciences, under NSF award \#1931555. It was also supported in part by funding from the Heliophysics Living With a Star Science Program, under NASA award \#NNX15AF39G.

\bibliographystyle{IEEEtran}
\bibliography{mybib}

\end{document}